\documentclass{sig-alternate-2013}

\usepackage{graphicx}
\usepackage{amsmath}
\usepackage{changepage}
\usepackage{float}
\usepackage{hyperref}
\usepackage{breakurl}
\usepackage{todonotes}
\usepackage{fancyvrb}
\usepackage{dsfont}
\usepackage{amssymb}
\usepackage{verbatim}
\usepackage{tikz,pgfplots}
\usepackage{indentfirst}
\usepackage{breqn}
\usepackage{pbox}
\usepackage{soul}
\usepackage{rotating}
\usepackage{slashbox}
\usepackage{footmisc}

\usetikzlibrary{bayesnet}

\usepackage{array}
\newcolumntype{L}[1]{>{\raggedright\let\newline\\\arraybackslash\hspace{0pt}}m{#1}}
\newcolumntype{C}[1]{>{\centering\let\newline\\\arraybackslash\hspace{0pt}}m{#1}}
\newcolumntype{R}[1]{>{\raggedleft\let\newline\\\arraybackslash\hspace{0pt}}m{#1}}

\renewcommand{\Re}{{\mathbb R}}
\newcommand{\x}{{\mathbf x}}

\newcommand{\m}{{\mathbf m}}
\newcommand{\cc}{{\mathbf c}}
\newcommand{\E}{{\mathds E}}
\newcommand{\e}{{\textbf{e}}}
\renewcommand{\L}{{\mathcal{L}}}

\DeclareMathOperator*{\argmax}{arg\,max}

\begin{document}
\title{Probabilistic Bag-Of-Hyperlinks Model for Entity  Linking }

\numberofauthors{5}
\author{
\alignauthor
Octavian-Eugen Ganea\\
       \affaddr{Dept. of Computer Science}\\
       \affaddr{ETH Zurich, Switzerland}\\
       \email{ganeao@inf.ethz.ch}
       \alignauthor
Marina Ganea\titlenote{Currently at Google Inc.}\\
       \affaddr{Dept. of Computer Science}\\
       \affaddr{ETH Zurich, Switzerland}\\
       \email{marinah@google.com}
       \alignauthor
Aurelien Lucchi\\
       \affaddr{Dept. of Computer Science}\\
       \affaddr{ETH Zurich, Switzerland}\\
       \email{alucchi@inf.ethz.ch}
\and 
Carsten Eickhoff\\
       \affaddr{Dept. of Computer Science}\\
       \affaddr{ETH Zurich, Switzerland}\\
       \email{ecarsten@inf.ethz.ch}         
       \alignauthor
Thomas Hofmann\\
       \affaddr{Dept. of Computer Science}\\
       \affaddr{ETH Zurich, Switzerland}\\
       \email{thomaho@inf.ethz.ch}
}

\maketitle
\begin{abstract}
Many fundamental problems in natural language processing rely on determining what entities appear in a given text. Commonly referenced as \textit{entity linking}, this step is a fundamental component of many NLP tasks such as text understanding, automatic summarization, semantic search or machine translation. Name ambiguity, word polysemy, context dependencies and a heavy-tailed distribution of entities contribute to the complexity of this problem.

We here propose a probabilistic approach that makes use of an effective graphical model to perform collective \textit{entity disambiguation}. Input mentions (i.e.,~linkable token spans) are disambiguated jointly across an entire document by combining a document-level prior of entity co-occurrences with local information captured from mentions and their surrounding context. The model is based on simple sufficient statistics extracted from data, thus relying on few parameters to be learned.

Our method does not require extensive feature engineering, nor an expensive training procedure. We use loopy belief propagation to perform approximate inference. The low complexity of our model makes this step sufficiently fast for real-time usage. We demonstrate the accuracy of our approach on a wide range of benchmark datasets, showing that it matches, and in many cases outperforms, existing state-of-the-art methods.

\end{abstract}



\section{Introduction}\label{sec:intro}

\begin{figure}[hb]
\centering
\includegraphics[width=0.45\textwidth]{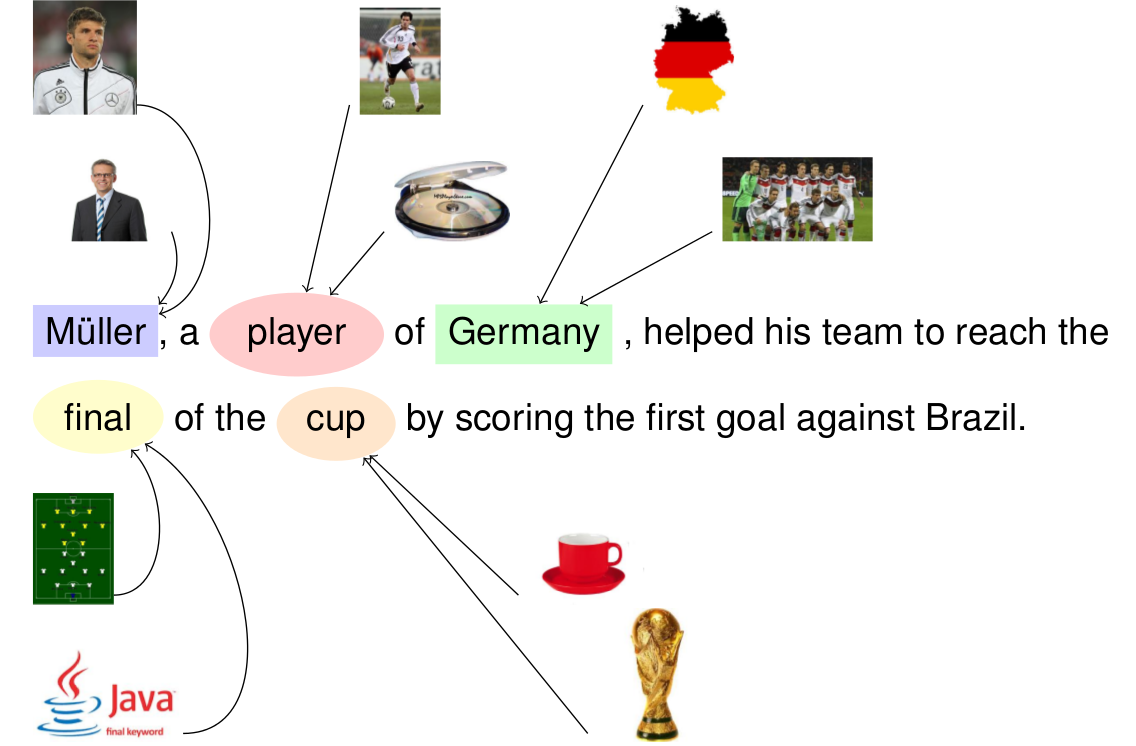}
\caption{An entity disambiguation problem showcasing five given mentions and their potential entity candidates.}\label{fig:example}
\end{figure}

Digital systems are producing increasing amounts of data every day. With daily global volumes of several terabytes of newly textual content, there is a growing need for automatic methods for text aggregation, summarization, and, eventually, semantic understanding. Entity linking is a key step towards these goals as it reveals the semantics of spans of text that refer to real-world entities. In practice, this is achieved by establishing a mapping between potentially ambiguous surface forms of entities and their canonical representations such as corresponding Wikipedia\footnote{\url{http://en.wikipedia.org/}} articles or Freebase\footnote{\url{https://www.freebase.com/}} entries. Figure \ref{fig:example} illustrates the difficulty of this task when dealing with real-world data. The main challenges arise from word ambiguities inherent to natural language: surface form \textit{synonymy}, i.e., different spans of text referring to the same entity, and \textit{homonymy}, i.e., the same name being shared by multiple entities.  

We here describe and evaluate a novel light-weight and fast alternative to heavy machine-learning approaches for document-level entity disambiguation with Wikipedia. Our model is primarily based on simple empirical statistics acquired from a training dataset and relies on a very small number of learned parameters. This has certain advantages like a very fast training procedure that can be applied to massive amounts of data, as well as a better understanding of the model compared to increasingly popular deep learning architectures (e.g., He et al.~\cite{he2013learning}). As a prerequisite, we assume that a given input set of mentions was already discovered via a \textit{mention detection} procedure\footnote{For example, using a named-entity recognition system. However, note that our approach is not restricted to named entities, but targets any Wikipedia entity.}. Our starting point is the natural assumption that each entity depends (i) on its mention, (ii) its neighboring local contextual words, and (iii) on other entities that appear in the same document.

In order to enforce these conditions, we rely on a conditional probabilistic model that consists of two parts: (1) the likelihood of a candidate entity given the referring token span and  its surrounding context, and (2) the prior joint distribution of the candidate entities corresponding to all the mentions in a document. Our model relies on the max-product algorithm to collectively infer entities for all mentions in a given document.

We further illustrate these modeling decisions. In the example depicted in Figure \ref{fig:example}, each highlighted mention constrains the set of possible entity candidates to a limited size set, yet leaves a significant level of ambiguity. However, there is one collective way of linking that is jointly consistent with all the chosen entities and supported by contextual cues. Intuitively, the related entities \texttt{Thomas\_M{\"u}ller} and \texttt{Germany\_national\_football\_team}
are likely to appear in the same document, especially in the presence of contextual words related to soccer, like ``team'' or ``goal''. 

Our main contributions are outlined below: \textbf{(1)} We employ rigorous probabilistic semantics for the \textit{entity disambiguation} problem by introducing a principled probabilistic graphical model that requires a simple and fast training procedure.  \textbf{(2)} At the core of our joint probabilistic model, we derive a minimal set of potential functions that proficiently explain statistics of observed training data. 
\textbf{(3)} Throughout a range of experiments performed on several standard  datasets using the Gerbil platform~\cite{gerbil}, we demonstrate competitive or state of the art quality compared to some of the best existing approaches. \textbf{(4)} Moreover, our training procedure is solely based on publicly available Wiki-pedia hyperlink statistics and the method does not require extensive hyperparameter tuning, nor feature engineering, making this paper a self-contained manual of implementing an entity disambiguation system from scratch. 

The remainder of this paper is structured as follows: Section \ref{sec:related} briefly discusses relevant entity linking literature. Section \ref{sec:model} formally introduces our probabilistic graphical model and details the initialization and learning procedure of the  model's parameters. Section \ref{sec:inference} describes the inference process used for collective entity resolution. Section \ref{sec:experiments} empirically demonstrates the merits of the proposed method on multiple standard collections of manually annotated documents. Finally, in Section \ref{sec:conclusion}, we conclude with a summary of our findings and an overview of ongoing and future work.

\section{Related Work}\label{sec:related}

There is a substantial body of existing work dedicated to the task of entity linking with Wikipedia (Wikification). We can identify four major paradigms of how this challenge is approached. 

\textit{Local models} consider the individual context of each entity mention in isolation in order to reduce the size of the decision space. In one of the early entity linking papers, Mihalcea and Csomai~\cite{mihalcea2007wikify} propose an entity disambiguation scheme based on similarity statistics between the mention context and the entity's Wikipedia page. Milne and Witten~\cite{milne2008learning} further refine their scheme with special focus on the mention detection step. Bunescu and Pasca~\cite{bunescu2006using} present a Wikipedia-driven approach, making use of manually created resources such as redirect and disambiguation pages. Dredze et al.~\cite{dredze2010entity} cast the entity linking task as a retrieval problem, treating mentions and their contexts as queries, and ranking candidate entities according to their likelihood of being referred to.

\textit{Global models} attempt to jointly disambiguate all mentions in a document based on the assumption that the underlying entities are correlated and consistent with the main topic of the document. While this approach tends to result in superior accuracy, the space of possible entity assignments grows combinatorially. As a consequence, many approaches in this group rely on approximate inference mechanisms. Cucerzan~\cite{cucerzan2007large} uses high-dimensional vector space representations of candidate entities and attempts to iteratively choose candidates that optimize the mutual proximity to existing candidates. Kulkarni et al.~\cite{kulkarni2009collective} exploit topical information about candidate entities and try to harmonize these topics across all assigned entities. Ratinov et al.~\cite{ratinov2011local} prune the list of entity mentions using support vector machines trained on a range of similarity and term overlap features between entity representations. Ferragina and Scaiella~\cite{ferragina2010tagme} focus on short documents such as tweets or search engine snippets. Based on evidence across all mentions, the authors employ a voting scheme for entity disambiguation. Cheng et al.~\cite{cheng2013relational} and Singh et al.~\cite{singh2013joint} describe models for jointly capturing the interdependence between the tasks of entity tagging, relation extraction and co-reference resolution. Similarly, Durrett and Klein~\cite{durrett2014joint} describe a graphical model for collectively addressing the tasks of named entity recognition, entity disambiguation and co-reference resolution. 

\textit{Graph-based models} establish relationships between candidate entities and mentions using structural models. For inference, various approaches are employed, ranging from densest graph estimation algorithms (Hoffart et al.~\cite{aida}) to graph traversal methods such as random graph walks (Guo and Barbosa~\cite{rel-rw}, Han et al.~\cite{han2011collective}). In a similar fashion, these techniques can be combined to enhance the quality of both entity linking and word sense disambiguation in a synergistic solution (Moro et al.~\cite{babelfy}). 

The above approaches are limited because they assume a single topic per document. Naturally, \textit{topic modelling} can be used for entity disambiguation by attempting to harmonize the individual distribution of latent topics across candidate entities. Houlsby and Ciaramita~\cite{houlsby2014scalable} and Pilz and Paa{\ss}~\cite{pilz2011names} rely on Latent Dirichlet Allocation (LDA) and compare the resulting topic distribution of the input document to the topic distributions of the disambiguated entities' Wikipedia pages. Han and Sun~\cite{han2012entity} propose a joint model of mention context compatibility and topic coherence, allowing them to simultaneously draw from both local (terms, mentions) as well as global (topic distributions) information. Kataria et al.~\cite{kataria2011entity} use a semi-supervised hierarchical LDA model based on a wide range of features extracted from Wikipedia pages and topic hierarchies.

In contrast to previous work on this problem, our method exploits co-occurrence statistics in a fully probabilistic manner using a graph-based model that addresses collective entity disambiguation. It combines a clean and light-weight probabilistic model with an elegant, real-time inference algorithm. An advantage over increasingly popular deep learning architectures for entity linking (e.g. Sun et al.~\cite{sun2015modeling}, He et al.~\cite{he2013learning}) is the speed of our training procedure that relies on count statistics from data and that learns only very few parameters. State-of-art accuracy is achieved without the need for special-purpose computational heuristics.

\section{Probabilistic Model}\label{sec:model}

In this section, we formally define the \textit{entity linking} task that we address in this work and describe our modeling approach in detail. 

\subsection{Problem Definition and Formulation}\label{ssec:probl-def}

Let $\mathcal{E}$ be a knowledge base (KB) of entities, $\mathcal{V}$ a finite dictionary of phrases or names and $\mathcal{C}$ a context representation. Formally, we seek a mapping $F: ( {\mathcal V}, {\mathcal C})^n \to \mathcal{E}^n$, that takes as input a sequence of linkable \textit{mentions} $\m=(m_1,\dots,m_n)$ along with their contexts $\cc=(c_1,\dots,c_n)$ and produces a joint entity assignment $\e=(e_1,\dots,e_n)$. Here $n$ refers to the number of linkable spans in a document. Our problem is also known as \textit{entity disambiguation} or \textit{link generation} in the literature. \footnote{Note that we do not address the issues of \textit{mention detection} or \textit{nil identification} in this work. Rather, our input is a document along with a fixed set of linkable mentions corresponding to existing KB entities.}

We can construct  such a mapping $F$ in a probabilistic approach, by learning a conditional probability model $p(\e | \m, \cc)$ from data and then employing (approximate) probabilistic inference in order to find the maximum a posteriori (MAP) assignment, hence: 
\begin{align}
F(\m,\cc)  := \argmax_{\e \in \mathcal{E}^n} p(\e | \m, \cc) \, .
\label{eq:posterior}
\end{align}
In the sequel, we describe how to estimate such a model from a corpus of entity-linked documents. Finally, we show in Section \ref{sec:inference} how to apply belief propagation (max-product) for approximate inference in this model.

\subsection{Maximum Entropy Models}

Assume a corpus of entity-linked documents is available. Specifically, we used the set of Wikipedia pages together with their respective Wiki hyperlinks. These hyperlinks are considered ground truth annotations, the mention being the linked span of text and the truth entity being the Wikipedia page it refers to. One can extract two kinds of basic statistics from such a corpus: First, counts of how often each entity was referred to by a specific name. Second, pairwise co-occurrence counts for entities in documents. Our fundamental conjecture is that most of the relevant information needed for entity disambiguation is contained in these counts, that they are \textit{sufficient statistics}. We thus request that our probability model reproduces these counts in expectation. As this alone typically yields an ill-defined problem, we follow the \textit{maximum entropy principle} of Jaynes~\cite{jaynes1982rationale}: Among the feasible set of distributions we favor the one with maximal entropy. 

Formally, let ${\mathcal D}$ be an entity-linked document collection. Ignoring mention contexts for now, we extract for each document $d \in {\mathcal D}$ a sequence of mentions $\m^{(d)}$ and their corresponding target entities $\e^{(d)}$, both of length $n^{(d)}$. Assuming exchangeability of random variables within these sequences, we reduce each $(\e,\m)$ to statistics (or \textit{features}) about mention-entity and entity-entity co-occurrence as follows:
\begin{align}
\phi_{e, m}(\e,\m) & := \sum_{i=1}^n \mathds{1}[e_i = e] \! \cdot \! \mathds{1}[m_i = m], \; \forall (e,m) \! \in \! \mathcal{E} \! \times \!\mathcal{V} 
\\
\psi_{\{e,e'\}}(\e) & := \sum_{i < j} \mathds{1}[\{ e_i, e_j\} =  \{e,e'\}], \;  \forall e, e' \in \mathcal{E}\,,
\end{align}
where $\mathds{1}[\cdot]$ is the indicator function. Note that we use the subscript notation $\{e,e'\}$ for $\psi$ to take into account the symmetry in $e, e'$ as well the fact that one may have $e=e'$. 

The document collection provides us with empirical estimates for the expectation of these statistics under an i.i.d.~sampling model for documents, namely the averages
\begin{align}
\phi_{e,m}({\cal D}) & := \frac 1{| \cal D|} \sum_{d \in {\cal D}} \phi_{e,m}(\e^{(d)},\m^{(d)})\,,\\
\psi_{\{e,e'\}}({\cal D}) & := \frac 1{| \cal D|} \sum_{d \in {\cal D}} \psi_{\{e,e'\}}(\e^{(d)}) \,.
\end{align}  

Note that in entity disambiguation, the mention sequence $\m$ is always considered given, while we seek to predict the corresponding entity sequence $\e$. It is thus not necessary to try to model the joint distribution $p(\e,\m)$, but sufficient to construct a conditional model $p(\e|\m)$. Following Berger et al.~\cite{berger1996maximum} this can be accomplished by taking the empirical distribution $p(\m|{\cal D})$ of mention sequences and combining it with a conditional model via $p(\e,\m) = p(\e|\m) \cdot p(\m |{ \cal D})$. We then require that:
\begin{align}
\label{eq:constraints}
\E_p[ \phi_{e,m} ] = \phi_{e,m}({\cal D}) \quad \text{and} \quad
\E_p[ \psi_{\{e,e'\}} ] = \psi_{\{e,e'\}}({\cal D}),
\end{align}
which yields $|\mathcal{E}| \cdot | \mathcal{V}| + \binom{| \mathcal{E} |}{2} + |\mathcal{E}|$ moment constraints on $p(\e|\m)$.

The maximum entropy distributions, fulfilling constraints as stated in Eq.~\eqref{eq:constraints} form a conditional exponential family for which $\phi(\cdot,\m)$ and $\psi(\cdot,\cdot)$ are sufficient statistics. We thus know that there are canonical parameters $\rho_{e,m}$ and $\lambda_{\{e,e'\}}$ (formally corresponding to Lagrange multipliers) such that the maximum entropy distribution can be written as 
\begin{align}
\label{eq:cond-exp-family}
p(\e|\m; \rho, \lambda)= \frac 1{Z(\m)} \exp \left[ \langle \rho, \phi(\e,\m) \rangle + \langle \lambda, \psi(\e) \rangle \right]
\end{align}
where $Z(\m)$ is the partition function 
\begin{align}
Z(\m) := \sum_{\e  \in \mathcal{E}^n} \exp \left[ \langle \rho, \phi(\e,\m) \rangle + \langle \lambda, \psi(\e) \rangle \right]\,.
\end{align}
Here we interpret $(e,m)$ and $\{e,e'\}$ as multi-indices and suggestively define the shorthands
\begin{align}
\langle \rho, \phi \rangle := \sum_{e,m} \rho_{e,m} \phi_{e,m}, 
\;\; \langle \lambda, \psi \rangle := \sum_{\{ e,e'\}} \lambda_{\{e,e'\}} \psi_{\{e,e'\}} \,.
\end{align}
Note that we can switch between the statistics view and the raw data view by observing that 
\begin{align}
\langle \rho, \phi(\e,\m) \rangle = \sum_{i=1}^n \rho_{e_i,m_i}, \quad 
\langle \lambda, \psi(\e) \rangle = \sum_{i<j} \lambda_{\{e_i,e_j\}} \,.
\label{eq:unrolled}
\end{align}
While the maximum entropy principle applied to our fundamental conjecture restricts the form of our model to a finite-dimensional exponential family, we need to investigate ways of finding the optimal or -- as we will see -- an approximately optimal distribution in this family. To that extent, we first re-interpret the obtained model as a factor graph model. 

\subsection{Markov Network and Factor Graph} 

Complementary to the maximum entropy estimation perspective, we want to present a view on our model in terms of probabilistic graphical models and factor graphs. Inspecting Eq.~\eqref{eq:cond-exp-family} and interpreting $\phi$ and $\psi$ as potential functions, we can recover a Markov network that makes conditional independence assumptions of the following type: an entity link $e_i$ and a mention $m_j$ with $i \neq j$ are independent, given $m_i$ and $\e_{-i}$, where $\e_{-i}$ denotes the set of entity variables in the document excluding $e_i$. This means that a mention $m_j$ only influences a variable $e_i$ through the intermediate variable $e_j$. However, the functional form in Eq.~\eqref{eq:cond-exp-family} goes beyond these conditional independences in that it limits the order of interaction among the variables. A variable $e_i$ interacts with neighbors in its Markov blanket through pairwise potentials. In terms of a \textit{factor graph} decomposition, $p(\e|\m)$ decomposes into functions of two arguments only, modeling pairwise interactions between entities on one hand, and between entities and their corresponding mentions on the other hand.

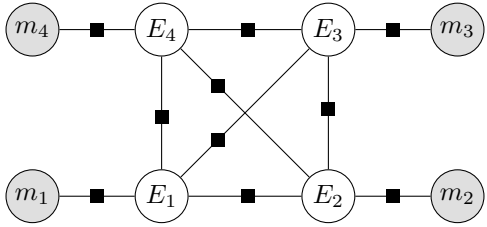
\begin{figure}
\centering{}
\begin{tikzpicture}
  \node[latent] (y1) {$E_1$};
  \node[latent, right=1.5cm of y1] (y2) {$E_2$};
  \node[latent, above=1.5cm of y2] (y3) {$E_3$};
  \node[latent, left=1.5cm of y3] (y4) {$E_4$};
  
  \node[obs, left =1 of y1] (x1) {$m_1$};
  \node[obs, right=1 of y2] (x2) {$m_2$};
  \node[obs, right=1 of y3] (x3) {$m_3$};
  \node[obs, left=1 of y4] (x4) {$m_4$};

  \factor[right=0.4 of x1] {x1-y1} {} {} {} ;
  \factor[left=0.4 of x2] {x2-y2} {} {} {} ;
  \factor[left=0.4 of x3] {x3-y3} {} {} {} ;
  \factor[right=0.4 of x4] {x4-y4} {} {} {} ;

  \edge[-] {x1}  {y1};
  \edge[-] {x2}  {y2};
  \edge[-] {x3}  {y3};
  \edge[-] {x4}  {y4};

\factor[right=0.7 of y1] {y1-y2} {} {}{};
\factor[right=0.7 of y4] {y3-y4} {} {}{};
 \factor[below=0.3 of y4, xshift=0.75cm] {y4-y2} {} {} {} ;
 \factor[above=0.3 of y1, xshift=0.75cm] {y1-y3} {} {} {} ;
\factor[below=0.6 of y3] {y2-y3} {} {}{};
\factor[above=0.6 of y1] {y2-y3} {} {}{};
 
  \edge[-] {y1}  {y2};
  \edge[-] {y1}  {y3};
  \edge[-] {y1}  {y4};
  \edge[-] {y2}  {y3};
  \edge[-] {y2}  {y4};
  \edge[-] {y3}  {y4};
\end{tikzpicture}
\caption{Proposed factor graph for a document with four mentions. Each mention node $m_i$ is paired with its corresponding entity node $E_i$, while all entity nodes are connected through entity-entity pair factors.}
\label{example-crf-el}
\end{figure}
We  emphasize the factor model view by rewriting \eqref{eq:cond-exp-family} as
\begin{align}
p(\e | \m; \rho, \lambda) \propto \prod_{i} \exp\left[ \rho_{e_i,m_i} \right] \cdot \prod_{i<j} \exp \left[ \lambda_{\{e_i,e_j\}} \right]
\label{eq:mrf}
\end{align}
where we think of $\rho$ and $\lambda$ as functions 
\begin{align}
& \rho: \mathcal{E} \times \mathcal{V} \to \Re, \quad (e,m) \mapsto \rho_{e,m} \nonumber \\
& \lambda: \mathcal{E} \cup \mathcal{E}^2 \to \Re, \quad \{e,e'\} \mapsto \lambda_{\{e,e'\}} \nonumber 
\end{align}
An example of a factor graph ($n=4$) is shown in Figure \ref{example-crf-el}. We will investigate in the sequel how the factor graph structure can be further exploited.

\subsection{(Pseudo--)Likelihood Maximization}\label{ssec:pseudolikelihood}

While the maximum entropy approach directly motivates the exponential form of Eq.~\eqref{eq:cond-exp-family} and is amenable to a plausible factor graph interpretation, it does not by itself suggest an efficient parameter fitting algorithm. As is known by convex duality, the optimal parameters can be obtained by maximizing the conditional likelihood of the model under the data,
\begin{align}
\L(\rho,\lambda; \mathcal{D}) = \sum_{d} \log\ p(\e^{(d)} | \m^{(d)}; \rho, \lambda)
\label{eq:likelihood}
\end{align}
However, specialized algorithms for maximum entropy estimation such as generalized iterative scaling \cite{darroch1972generalized} are known to be slow, whereas gradient-based methods require the computation of gradients of $\L$, which involves evaluating expectations with regard to the model, since     
\begin{align}
\nabla_\rho \log Z(\m) = \E_p \phi(\e,\m), \;\;\nabla_\lambda \log Z(\m) = \E_p \psi(\e) \,.
\end{align} 
The exact inference problem of computing these model expectations, however, is not generally tractable due to the pairwise couplings through the $\psi$-statistics.\\

As an alternative to maximizing the  likelihood in Eq.~\eqref{eq:likelihood}, we have investigated an approximation known as, pseudo-likelihood maximization \cite{mccallum2009,murphy2006}. Its main benefits are low computational complexity, simplicity and practical success. Switching to the Markov network view, the pseudo-likelihood estimator predicts each variable conditioned on the value of all variables in its Markov blanket. The latter consists of the  minimal set of variables that renders a variable conditionally independent of everything else. In our case the Markov blanket consists of all variables that share a factor with a given variable. Consequently, the Markov blanket of $e_i$ is ${\cal N}(e_i) := (m_i,\e_{-i})$. The posterior is then approximated in the pseudo-likelihood approach as:
\begin{align}
\tilde p(\e | \m; \rho, \lambda) := \prod_{i=1}^n p(e_i | {\cal N}(e_i); \rho, \lambda )\,,
\end{align}
which results in the tractable log-likelihood function
\begin{align} 
\label{eq:pseudo}
\tilde \L(\rho,\lambda; \mathcal{D}) :=
\sum_{d \in \mathcal{D}} \sum_{i=1}^{n^{(d)}} \log p(e^{(d)}_i | {\cal N}(e^{(d)}_i); \rho, \lambda )\,.
\end{align} 

Introducing additional $L_2$-norm penalties $\gamma (\| \lambda\|^2_2 + \| \rho\|_2^2)$ to further regularize $\tilde \L$, we have utilized parallel stochastic gradient descent (SGD)~\cite{hogwild} with sparse updates to learn parameters $\rho, \lambda$. From a practical perspective, we only keep for each token span $m$ parameters $\rho_{e,m}$ for the most frequently observed entities $e$. Moreover, we only use  $\lambda_{\{e,e'\}}$ for entity pairs $(e,e')$ that co-occurred together a sufficient number of times in the collection $\mathcal{D}$.\footnote{For the Wikipedia collection, even after these pruning steps, we ended up with more than 50 million parameters in total.} As we will discuss in more detail in Section \ref{sec:experiments}, our experimental findings suggest this brute-force learning approach to be somewhat ineffective, which has motivated us to develop simpler, yet more effective plug-in estimators as described below. 

\subsection{Bethe Approximation} \label{ssec:unnormalized}

The major computational difficulty with our model lies in the pairwise couplings between entities and the fact that these couplings are dense: The Markov dependency graph between different entity links in a document is always a complete graph. Let us consider what would happen, if the dependency structure were loop-free, i.e.,~it would form a tree. Then we could rewrite the prior probability in terms of marginal distributions in the so-called \textit{Bethe form}. Encoding the tree structure in a symmetric relation ${\cal T}$, we would get  
\begin{align}
p(\e) = \frac{ \prod_{ \{i,j\} \in {\cal T} }  p(e_i,e_j) }{
   \prod_{i = 1}^n p(e_i)^{d_i - 1} } , \quad d_i := |\{ j: \{i,j\} \in {\cal T}\}|
   \,.
\label{eq:bethe-exact}
\end{align}
The Bethe approximation \cite{bethe1} pursues the idea of using the above representation as an unnormalized approximation for $p(\e)$, even when the Markov network has cycles. How does this relate to the exponential form in Eq.~\eqref{eq:cond-exp-family}? By simple pattern matching, we see that if we choose 
\begin{align}
\lambda_{\{e,e'\}} &= \log\left(\frac{p(e,e')}{p(e) \,p(e')}\right),\quad  \forall e, e' \in \mathcal{E}
\label{eq:lambda-bethe}
\end{align}
we can apply Eq.~\eqref{eq:bethe-exact} to get an approximate distribution
\begin{align}
\label{eqn:lambda-matching}
\begin{split}
\bar p(\e) & \propto \frac{ \prod_{i<j}  p(e_i,e_j) }{\prod_{i=1}^n p(e_i)^{n-2} } =  \prod_{i=1}^n p(e_i) \prod_{i<j} \frac{p(e_i,e_j)}{p(e_i)\, p(e_j)} \\
&=  \exp\left[ \sum_{i} \log p(e_i) + \sum_{i<j}  \lambda_{\{e_i,e_j\}} \right] \,,
\end{split}
\end{align}
where we see the same exponential form in $\lambda$ appearing as in Eq.~\eqref{eq:unrolled}. We complete this argument by observing that with 
\begin{align}
\rho_{e,m} = \log p(e) + \log p(m|e) 
\label{eq:rho}
\end{align}
we obtain a representation of a joint distribution that exactly matches the form in Eq.~\eqref{eq:cond-exp-family}.\\

What have we gained so far? We started from the desire of constructing a model that would agree with the observed data on the co-occurrence probabilities of token spans and their linked entities as well as on the co-link probability of entity pairs within a document. This has led to the conditional exponential family in Eq.~\eqref{eq:cond-exp-family}. We have then proposed pseudo-likelihood maximization as a way to arrive at a tractable learning algorithm to try to fit the massive amount of parameters $\rho$ and $\lambda$. Alternatively, we have now seen that a Bethe approximation of the joint prior $p(\e)$ yields a conditional distribution $p(\e|\m)$ that (i) is a member of the same exponential family, (ii) has explicit formulas for how to choose the parameters from pairwise marginals, and (iii) would be exact in the case of a dependency tree. We claim that the benefits of computational simplicity together with the correctness guarantee for non-dense dependency networks outweighs the approximation loss, relative to the model with the best generalization performance within the conditional exponential family. In order to close the suboptimality gap further, we suggest some important refinements below.

\subsection{Parameter Calibration} \label{ssec:scaling-len}

With the previous suggestion, one issue comes into play: The total contribution coming from the pairwise interactions between entities will scale with $\binom n 2$, while the entity--mention compatibility contributions will scale with $n$, the total number of mentions. This is a direct observation of the number of terms contributing to the sums in \eqref{eq:unrolled}. However, for practical reasons, it is somewhat implausible that, as $n$ grows, the prior $p(\e)$ should dominate and the contribution of the likelihood term should vanish. The model is not well-calibrated with regard to  $n$. 

We propose to correct for this effect by adding a normalization factor to the $\lambda$-parameters by replacing \eqref{eq:lambda-bethe} with:
\begin{align}
\lambda^n_{e,e'} = \frac{2}{n-1} \log\left(\frac{p(e,e')}{p(e)\cdot p(e')}\right), \quad  \forall e,e' \in \mathcal{E}
\label{eq:bethe-normalized-params}
\end{align}
where now these parameters scale inversely with $n$, the number of entity links in a document, making the corresponding sum in Eq.~\eqref{eq:cond-exp-family} scale with $n$. With this simple change, a substantial accuracy improvement was observed empirically, the details of which are reported in our experiments.\\

The re-calibration in Eq.~\eqref{eq:bethe-normalized-params} can also be justified by the following combinatorial argument: 
For a given set ${\mathbf Y}$ of random variables, define an ${\mathbf Y}$-cycle as a graph containing as nodes all variables in ${\mathbf Y}$, each with degree exactly 2, connected in a single cycle. Let $\Xi$ be the set enumerating all possible ${\mathbf Y}$-cycles. Then, $|\Xi| = (n-1)!$, where $n$ is the size of ${\mathbf Y}$.

In our case, if the entity variables $\e$ per document would have formed a cycle of length $n$ instead of a complete subgraph, the Bethe approximation would have been written as:
\begin{align}
\bar p_\pi(\e) \propto \frac{\prod_{(i,j) \in E(\pi)} p(e_i,e_j)}{\prod_i p(e_i) }, \quad \forall \pi \in \Xi
\end{align}
where $E(\pi)$ is the set of edges of the $\e$-cycle $\pi$. However, as we do not desire to further constrain our graph with additional independence assumptions, we propose to approximate the joint prior $p(\e)$ by the average of the Bethe approximation of all possible $\pi$, that is 
\begin{align}
\log \bar p(\e) \approx \frac{1}{|\Xi|} \sum_{\pi \in \Xi} \log \bar p_\pi(\e) \,.
\end{align}
Since each pair $(e_i,e_j)$ would appear in exactly $2(n-2)!$ $\e$-cycles, one can derive the final approximation:
\begin{align}
\bar p(\e) \approx \frac{\prod_{i < j} p(e_i,e_j)^{\frac{2}{n-1}}}{\prod_i p(e_i) }\,.
\label{eq:average-bethe}
\end{align}
Distributing marginal probabilities over the parameters starting from Eq.~\eqref{eq:average-bethe} and applying a similar argument as in Eq.~\eqref{eqn:lambda-matching} results in the assignment given by Eq.~\eqref{eq:bethe-normalized-params}. While the above line of argument is not a strict mathematical derivation, we believe this to shed further light on the empirically observed effectiveness of the parameter re-scaling. 

\subsection{Integrating Context}  \label{ssec:smoothed}

The model that we have discussed so far does not consider the local context of a mention. This is a powerful source of information that a competitive entity linking system should utilize. For example, words like ``computer'', ``company'' or ``device''  are more likely to appear near references of the entity \texttt{Apple\_Inc.} than of the entity \texttt{Apple\_fruit}.  We demonstrate in this section how this integration can be easily done in a principled way on top of the current probabilistic model. This showcases the extensibility of our approach. Enhancing our model with additional knowledge such as entity categories or word co-reference can also be done in a rigorous way, so we hope that this provides a template for future extensions. 

As stated in Section~\ref{ssec:probl-def}, for each mention $m_i$ in a document, we maintain a context representation $c_i$ consisting of the bag of words surrounding the mention within a window of length $K$\footnote{Throughout our experiments, we used a context window of size $K = 100$, intuitively chosen and without extensive validation.}. Hence, $c_i$ can be viewed as an additional random variable with an observed outcome.  At this stage, we make additional reasonable independence assumptions that increase  tractability of our model. First, we assume that,  knowing the identity of the linked entity $e_i$, the mention token span $m_i$ is just the surface form of the entity, so it brings no additional information for the generative process describing the surrounding context $c_i$. Formally, this means that $m_i$ and $c_i$ are conditionally independent given $e_i$. Consequently, we obtain a factorial expression for the joint model
\begin{align}
p(\e, \m, \cc) = p(\e) p(\m, \cc| \e) = p(\e) \prod_{i=1}^n p(m_i|e_i) p(c_i|e_i)
\label{eq:context-mention}
\end{align}
This is a simple extension of the previous factor graph that includes context variables. Second, we assume conditional independence of the words in $c_i$ given an entity $e_i$ which let us factorize the context probabilities as 
\begin{align}
p(c_i | e_i) = \prod_{w_j \in c_i} p(w_j | e_i)\,.
\label{eq:context-factor}
\end{align}
Note that this assumption is commonly made in models using bag-of-word representations or na\"{\i}ve Bayes classifiers.

While this completes the argument from a joint model point of view, we need to consider one more aspect for the conditional distribution $p(\e|\m,\cc)$ that we are interested in. If we cannot afford (computationally as well as with regard to training data size) a full-blown discriminative learning approach, then how do we balance the relative influence of the context $c_i$ and the mention token span $m_i$ on $e_i$? For instance, the effect of $c_i$ will depend on the chosen window size $K$, which is not realistic.  

To address this issue, we resort to a hybrid approach, where, in the spirit of the Bethe approximation, we continue to express our model in terms of simple marginal distributions that can be easily estimated independently from data, yet that allow for a small number of parameters (in our case ``small'' equals $2$) to be chosen to optimize the conditional log-likelihood $p(\e|\m,\cc)$. We thus introduce weights $\zeta$ and $\tau$ that control the importance of the context factors and, respectively, of the entity-entity interaction factors. Putting equations~\eqref{eq:rho},~\eqref{eq:bethe-normalized-params},~\eqref{eq:context-mention} and~\eqref{eq:context-factor} together, we arrive at the final model that will be subsequently referred to as the \textit{PBoH} model (\textbf{P}robabilistic \textbf{B}ag \textbf{o}f \textbf{H}yperlinks):

\begin{align}
\log p(\e|\m, \cc) = & \sum_{i=1}^n \left( \log p(e_i|m_i )+ \zeta  \sum_{w_j \in c_i} \log p(w_j | e_i) \right) \nonumber \\
&  + \frac{2 \tau}{n-1}  \sum_{i < j} \log\left(\frac{p(e_i,e_j)}{p(e_i) \, p(e_j)}\right) \!+ \!\text{const} \,.
\label{eq:c2}
\end{align}
Here we used the identity $p(m|e) p(e) = p(e|m) p(m)$ and absorbed all $\log p(m)$ terms in the constant.  We use grid-search on a validation set for the remaining problem of optimizing over the parameters $\zeta,\tau$. Details are provided in section \ref{sec:experiments}.

\subsection{Smoothing Empirical Probabilities}

In order to estimate the probabilities involved in Eq.~\eqref{eq:c2}, we rely on an entity annotated corpus of text documents, e.g.,~Wikipedia Web pages together with their hyperlinks which we view as ground truth annotations. From this corpus, we derive empirical probabilities for a name-to-entity dictionary $\hat p(m|e)$ based on counting how many times an entity appeared referenced by a given name\footnote{In our implementation we summed the mention-entity counts from Wikipedia hyperlinks with the Crosswikis counts~\cite{spitkovsky2012lrec}}. We also compute the pairwise probabilities $\hat{p}(e,e')$ obtained by counting the pairwise co-occurrence of entities $e$ and $e'$ within the same document. Similarly, we obtained empirical values for the marginals $\hat{p}(e) = \sum_{e'} \hat{p}(e,e')$ and for the context word-entity statistics $\hat{p}(w|e)$.

In the absence of huge amounts of data, estimating such probabilities from counts is subject to sparsity. For instance, in our statistics, there are 8 times more distinct pairs of entities that co-occur in at most 3 Wikipedia documents compared to the total number of distinct pairs of entities that appear together in at least 4 documents. Thus, it is expected that the heavy tail of infrequent pairs of entities will have a strong impact on the accuracy of our system. 

Traditionally, various smoothing techniques are employed to address sparsity issues arising commonly in areas such as natural language processing. Out of the wealth of methods, we decided to use the absolute discounting smoothing technique~\cite{zhai2004study} that involves interpolation of higher and lower order (backoff) models. In our case, whenever insufficient data is available for a pair of entities $(e,e')$, we assume the two entities are drawn from independent distributions. Thus, if we denote by $N(e,e')$ the total number of corpus documents that link both $e$ and $e'$, and by $N_{ep}$ the total number of pairs of entities referenced in each document, then the final formula for the smoothed entity pairwise probabilities is:
\begin{align}
\widetilde{p}(e,e') = \frac{\max(N(e,e') - \delta, 0)}{N_{ep}} + (1 - \mu_e) \hat{p}(e) \hat{p}(e')
\end{align}
where $\delta \in [0,1]$ is a fixed discount and $\mu_e$ is a constant that assures that $\sum_{e} \sum_{e'} \widetilde{p}(e,e') = 1$. $\delta$ was set by performing a coarse grid search on a validation set. The best $\delta$ value was found to be 0.5.

The word-entity empirical probabilities $\hat{p}(w|e)$ were computed based on the Wikipedia corpus by counting the frequency with which word $w$ appears in the context windows of size K around the hyperlinks pointing to $e$. In order to avoid memory explosion, we only considered the entity-words pairs for which these counts are at least 3. 
These empirical estimates are also sparse, so we used absolute discounting smoothing for their correction by backing off to the unbiased estimates $\hat{p}(w)$. The latter can be much more accurately estimated from any text corpus. Finally, we obtain:
\begin{align}
\widetilde{p}(w | e) = \frac{\max(N(w,e) - \xi, 0)}{N_{wp}} + (1 - \mu_w) \hat{p}(w)\,.
\end{align}
Again $\xi \in [0,1]$ was optimized by grid search to be 0.5.

\section{Inference}\label{sec:inference}

After introducing our model and showing how to train it in the previous
section, we now explain the inference process used for prediction.


\subsection{Candidate Selection}  \label{ssec:candidates}

At test time, for each mention to be disambiguated, we first select a set of potential candidates by considering the top $R$ ranked entities based on the local mention-entity probability dictionary $\hat{p}(e|m)$. We found $R=64$ to be a good compromise between efficiency and accuracy loss. Second, we want to keep the average number of candidates per mention as small as possible in order to reduce the running time which is quadratic in this number (see the next section for details). Consequently, we further limit the number of candidates per mention by keeping only the top 10 entity candidates re-ranked by the local mention-context-entity compatibility defined as
\begin{align}
\log p(e_i|m_i, c_i) = \log p(e_i | m_i) + \zeta \sum_{w_j \in c_i} \log p(w_j | e_i) + \!\text{const} \,.
\label{eq:local-all} 
\end{align}
These pruning heuristics result in a significantly improved running time at an insignificant accuracy loss.

If the given mention is not found in our map $\hat{p}(e|m)$, we try to replace it by the closest name in this dictionary. Such a name is picked only if the Jaccard distance between the set of letter trigrams of these two strings is smaller than a threshold that we empirically picked as 0.5. Otherwise, the mention is not linked at all.

\subsection{Belief Propagation}

Collectively disambiguating all mentions in a text involves iterating through an exponential number of possible entity resolutions. Exact inference in general graphical models is NP-hard, therefore approximations are employed. We propose solving the inference problem through the \textit{loopy belief propagation} (LBP)~\cite{loopy} technique, using the max-product algorithm that approximates the MAP solution in a run-time polynomial in $n$, the number of input mentions. For the sake of brevity, we only present the algorithm for the maximum entropy model described by Eq.~\eqref{eq:cond-exp-family}; A similar approach was used for the enhanced PBoH model given by Eq.~\eqref{eq:c2}.

Our proposed graphical model is a fully connected graph where each node corresponds to an entity random variable. Unary potentials $\exp(\rho_{m,e})$ model the entity-mention compatibility, while pairwise potentials $\exp(\lambda_{\{e,e'\}})$ express enti\-ty-entity correlations. For the posterior in Eq.~\eqref{eq:cond-exp-family}, one can derive the update equation of the logarithmic message that is sent in round $t + 1$ from entity random variable $E_i$ to the outcome $e_j$ of the entity random variable $E_j$ :
\begin{align}
m_{E_i \rightarrow E_j}^{t+1}&(e_j) =  \\
& \max_{e_i} \left( \rho_{e_i, m_i} \!+\! \lambda_{\{e_i, e_j\}} \!+ \! \sum \limits_{1 \leq k \leq n; k \neq j} m_{E_k \rightarrow E_i}^t(e_i) \right) \nonumber
\end{align}
Note that, for simplicity, we skip the factor graph framework and send messages directly between each pair of entity variables. This is equivalent to the original BP framework.

We chose to update messages synchronously: in each round $t$, each two entity nodes $E_i$ and $E_j$ exchange messages. This is done until convergence or until an allowed maximum number of iterations (15 in our experiments) is reached. The convergence criterion is:
\begin{align}
\max_{1 \leq i, j \leq n; e_j \in \mathcal{E}} \vert m_{E_i \rightarrow E_j}^{t+1}(e_j) - m_{E_i \rightarrow E_j}^t(e_j) \vert \le \epsilon
\end{align}
where $\epsilon = 10^{-5}$. This setting was sufficient in most of the cases to reach convergence. 

In the end, the final entity assignment is determined by:
\begin{align}
e_i^* = \arg\max_{e_i} \left( \rho_{e_i, m_i} + \sum \limits_{1 \leq k \leq n} m_{E_k \rightarrow E_i}^t(e_i) \right)
\end{align}

The complexity of the belief propagation algorithm is, in our case, $O(n^2 \cdot r^2)$, with $n$ being the number of mentions in a document and $r$ being the average number of candidate entities per mention (10 in our case). More details regarding the run-time and convergence of the loopy BP algorithm can be found in Section \ref{sec:experiments}.

\section{Experiments }\label{sec:experiments}


\begin{table}
\begin{tabular}{|c|c|c|}
\hline
Dataset &  \# non-NIL mentions & \# documents \\
\hline
AIDA test A & 4791 & 216 \\
AIDA test B & 4485 & 231 \\
MSNBC & 656 & 20 \\
AQUAINT & 727 & 50 \\
ACE04 & 257 & 35 \\
\hline
\end{tabular}
\caption{Statistics on some of the used datasets}
\label{Ta:datastats}
\end{table}

\begin{table*}[!ht]
\scriptsize
\centering

\begin{tabular}{@{} l|c| c| c| c| c| c| c| c| c| c| c| c| c| c| @{}}

\multicolumn{1}{p{1cm}}{\textbf{F1@MI  F1@MA}} & \multicolumn{1}{c}{\begin{turn}{90} ACE2004\end{turn}} & \multicolumn{1}{c}{\begin{turn}{90} AIDA/CoNLL-Complete \end{turn}} & \multicolumn{1}{c}{\begin{turn}{90} AIDA/CoNLL-Test A \end{turn}} & \multicolumn{1}{c}{\begin{turn}{90} AIDA/CoNLL-Test B \end{turn}} & \multicolumn{1}{c}{\begin{turn}{90} AIDA/CoNLL-Training \end{turn}} & \multicolumn{1}{c}{\begin{turn}{90} AQUAINT \end{turn}} & \multicolumn{1}{c}{\begin{turn}{90} DBpediaSpotlight \end{turn}} & \multicolumn{1}{c}{\begin{turn}{90} IITB \end{turn}} & \multicolumn{1}{c}{\begin{turn}{90} KORE50 \end{turn}} & \multicolumn{1}{c}{\begin{turn}{90} Microposts2014-Test \end{turn}} & \multicolumn{1}{c}{\begin{turn}{90} Microposts2014-Train \end{turn}} & \multicolumn{1}{c}{\begin{turn}{90} MSNBC \end{turn}} & \multicolumn{1}{c}{\begin{turn}{90} N3-Reuters-128 \end{turn}} & \multicolumn{1}{c}{\begin{turn}{90} N3-RSS-500 \end{turn}} \\

\hline
\hline

AGDISTIS &  \multicolumn{1}{p{0.6cm}|}{65.83 77.63} &  \multicolumn{1}{p{0.6cm}|}{60.27 56.97} &  \multicolumn{1}{p{0.6cm}|}{59.06 53.36} &  \multicolumn{1}{p{0.6cm}|}{58.32 58.03} &  \multicolumn{1}{p{0.6cm}|}{61.05 57.53} &  \multicolumn{1}{p{0.6cm}|}{60.10 58.62} &  \multicolumn{1}{p{0.6cm}|}{36.61 33.25} &  \multicolumn{1}{p{0.6cm}|}{41.23 43.38} &  \multicolumn{1}{p{0.6cm}|}{34.16 30.20} &  \multicolumn{1}{p{0.6cm}|}{42.43 61.08} &  \multicolumn{1}{p{0.6cm}|}{50.39 62.87} &  \multicolumn{1}{p{0.6cm}|}{75.42 73.82} &  \multicolumn{1}{p{0.6cm}|}{\textcolor{blue}{67.95 75.52}} &  \multicolumn{1}{p{0.6cm}|}{59.88 70.80}  \\
\hline
Babelfy &  \multicolumn{1}{p{0.6cm}|}{63.20 76.71} &  \multicolumn{1}{p{0.6cm}|}{78.00 73.81} &  \multicolumn{1}{p{0.6cm}|}{75.77 71.26} &  \multicolumn{1}{p{0.6cm}|}{80.36 74.52} &  \multicolumn{1}{p{0.6cm}|}{78.01 74.22} &  \multicolumn{1}{p{0.6cm}|}{72.27 73.23} &  \multicolumn{1}{p{0.6cm}|}{51.05 51.97} &  \multicolumn{1}{p{0.6cm}|}{57.13 55.36} &  \multicolumn{1}{p{0.6cm}|}{\textcolor{red}{73.12 69.77}} &  \multicolumn{1}{p{0.6cm}|}{47.20 62.11} &  \multicolumn{1}{p{0.6cm}|}{50.60 61.02} &  \multicolumn{1}{p{0.6cm}|}{78.17 75.73} &  \multicolumn{1}{p{0.6cm}|}{58.61 59.87} &  \multicolumn{1}{p{0.6cm}|}{69.17 76.00}  \\
\hline

DBpedia Spotlight &  \multicolumn{1}{p{0.6cm}|}{70.38	 80.02 } &  \multicolumn{1}{p{0.6cm}|}{58.84	60.59 } &  \multicolumn{1}{p{0.6cm}|}{54.90	 54.11 } &  \multicolumn{1}{p{0.6cm}|}{57.69 61.34 } &  \multicolumn{1}{p{0.6cm}|}{60.04	 62.23 } &  \multicolumn{1}{p{0.6cm}|}{74.03 73.13 } &  \multicolumn{1}{p{0.6cm}|}{69.27 67.23 } &  \multicolumn{1}{p{0.6cm}|}{\textcolor{blue}{65.44	 62.81} } &  \multicolumn{1}{p{0.6cm}|}{37.59	 32.90 } &  \multicolumn{1}{p{0.6cm}|}{56.43 71.63 } &  \multicolumn{1}{p{0.6cm}|}{56.26 67.99 } &  \multicolumn{1}{p{0.6cm}|}{69.27	 69.82 } &  \multicolumn{1}{p{0.6cm}|}{56.44 58.77 } &  \multicolumn{1}{p{0.6cm}|}{57.63 65.03 }  \\
\hline

Dexter &   \multicolumn{1}{p{0.6cm}|}{18.72 16.97 } &  \multicolumn{1}{p{0.6cm}|}{48.46 45.29 } &  \multicolumn{1}{p{0.6cm}|}{ 45.44 42.17} &  \multicolumn{1}{p{0.6cm}|}{48.59	 	46.20 } &  \multicolumn{1}{p{0.6cm}|}{49.25 45.85 } &  \multicolumn{1}{p{0.6cm}|}{38.28 38.15 } &  \multicolumn{1}{p{0.6cm}|}{26.70 22.75} &  \multicolumn{1}{p{0.6cm}|}{28.53	 28.48 } &  \multicolumn{1}{p{0.6cm}|}{17.20 12.54 } &  \multicolumn{1}{p{0.6cm}|}{31.27 44.02 } &  \multicolumn{1}{p{0.6cm}|}{35.21	 42.07 } &  \multicolumn{1}{p{0.6cm}|}{36.86 39.42 } &  \multicolumn{1}{p{0.6cm}|}{32.74 31.85 } &  \multicolumn{1}{p{0.6cm}|}{31.11 33.55 }    \\
\hline

Entityclassifier.eu &  \multicolumn{1}{p{0.6cm}|}{12.74  12.3} &  \multicolumn{1}{p{0.6cm}|}{46.6 42.86 } &  \multicolumn{1}{p{0.6cm}|}{44.13 42.36} &  \multicolumn{1}{p{0.6cm}|}{44.02 41.31} &  \multicolumn{1}{p{0.6cm}|}{47.83 43.36} &  \multicolumn{1}{p{0.6cm}|}{21.67  19.59} &  \multicolumn{1}{p{0.6cm}|}{22.59 18.0} &  \multicolumn{1}{p{0.6cm}|}{18.46 19.54} &  \multicolumn{1}{p{0.6cm}|}{27.97  25.2} &  \multicolumn{1}{p{0.6cm}|}{29.12	 39.53} &  \multicolumn{1}{p{0.6cm}|}{32.69 38.41} &  \multicolumn{1}{p{0.6cm}|}{41.24 40.3} &  \multicolumn{1}{p{0.6cm}|}{28.4 24.84} &  \multicolumn{1}{p{0.6cm}|}{21.77 22.2}    \\
\hline

Kea &  \multicolumn{1}{p{0.6cm}|}{80.08  87.57} &  \multicolumn{1}{p{0.6cm}|}{73.39  73.26} &  \multicolumn{1}{p{0.6cm}|}{70.9 67.91 } &  \multicolumn{1}{p{0.6cm}|}{72.64  73.31} &  \multicolumn{1}{p{0.6cm}|}{74.22 74.47 } &  \multicolumn{1}{p{0.6cm}|}{\textcolor{blue}{81.84 81.27} } &  \multicolumn{1}{p{0.6cm}|}{\textcolor{blue}{73.63  76.60}} &  \multicolumn{1}{p{0.6cm}|}{\textcolor{red}{72.03 70.52} } &  \multicolumn{1}{p{0.6cm}|}{57.95  53.17} &  \multicolumn{1}{p{0.6cm}|}{\textcolor{blue}{63.4 76.54}} &  \multicolumn{1}{p{0.6cm}|}{\textcolor{blue}{64.67  74.32}} &  \multicolumn{1}{p{0.6cm}|}{\textcolor{blue}{85.49 87.4}} &  \multicolumn{1}{p{0.6cm}|}{63.2 64.45} &  \multicolumn{1}{p{0.6cm}|}{69.29 75.93}     \\
\hline

NERD-ML &  \multicolumn{1}{p{0.6cm}|}{54.89 72.22} &  \multicolumn{1}{p{0.6cm}|}{54.62 52.35} &  \multicolumn{1}{p{0.6cm}|}{52.85 49.6} &  \multicolumn{1}{p{0.6cm}|}{52.59 51.34} &  \multicolumn{1}{p{0.6cm}|}{55.55  53.23} &  \multicolumn{1}{p{0.6cm}|}{49.68 46.06} &  \multicolumn{1}{p{0.6cm}|}{46.8 45.59} &  \multicolumn{1}{p{0.6cm}|}{51.08 49.91} &  \multicolumn{1}{p{0.6cm}|}{29.96  24.75} &  \multicolumn{1}{p{0.6cm}|}{38.65 57.91} &  \multicolumn{1}{p{0.6cm}|}{39.83 53.74} &  \multicolumn{1}{p{0.6cm}|}{64.03 67.28} &  \multicolumn{1}{p{0.6cm}|}{54.96 62.9} &  \multicolumn{1}{p{0.6cm}|}{61.22  67.3}  \\
\hline

TagMe 2 &  \multicolumn{1}{p{0.6cm}|}{\textcolor{blue}{81.93 89.09} } &  \multicolumn{1}{p{0.6cm}|}{72.07 71.19} &  \multicolumn{1}{p{0.6cm}|}{69.07 66.5} &  \multicolumn{1}{p{0.6cm}|}{70.62 70.38} &  \multicolumn{1}{p{0.6cm}|}{73.2  72.45} &  \multicolumn{1}{p{0.6cm}|}{76.27  75.12} &  \multicolumn{1}{p{0.6cm}|}{63.31 65.1 } &  \multicolumn{1}{p{0.6cm}|}{57.23  55.8} &  \multicolumn{1}{p{0.6cm}|}{57.34 54.67} &  \multicolumn{1}{p{0.6cm}|}{56.81 71.66 } &  \multicolumn{1}{p{0.6cm}|}{59.14  70.45} &  \multicolumn{1}{p{0.6cm}|}{75.96  77.05} &  \multicolumn{1}{p{0.6cm}|}{59.32 67.55} &  \multicolumn{1}{p{0.6cm}|}{\textcolor{red}{78.05 83.2 }}  \\
\hline

WAT &  \multicolumn{1}{p{0.6cm}|}{80.0 86.49} &  \multicolumn{1}{p{0.6cm}|}{\textcolor{blue}{83.82 83.59}} &  \multicolumn{1}{p{0.6cm}|}{\textcolor{blue}{81.82 80.25}} &  \multicolumn{1}{p{0.6cm}|}{\textcolor{blue}{84.34 84.12}} &  \multicolumn{1}{p{0.6cm}|}{\textcolor{blue}{84.21 84.22}  } &  \multicolumn{1}{p{0.6cm}|}{76.82 77.64 } &  \multicolumn{1}{p{0.6cm}|}{65.18  68.24 } &  \multicolumn{1}{p{0.6cm}|}{61.14  59.36} &  \multicolumn{1}{p{0.6cm}|}{58.99  53.13} &  \multicolumn{1}{p{0.6cm}|}{59.56 73.89 } &  \multicolumn{1}{p{0.6cm}|}{61.96  72.65 } &  \multicolumn{1}{p{0.6cm}|}{77.72  79.08} &  \multicolumn{1}{p{0.6cm}|}{64.38 65.81 } &  \multicolumn{1}{p{0.6cm}|}{68.21  76.0}  \\
\hline

Wikipedia Miner &  \multicolumn{1}{p{0.6cm}|}{77.14  86.36} &  \multicolumn{1}{p{0.6cm}|}{64.72  66.17 } &  \multicolumn{1}{p{0.6cm}|}{61.65  61.67} &  \multicolumn{1}{p{0.6cm}|}{60.71  63.19} &  \multicolumn{1}{p{0.6cm}|}{66.48  67.93 } &  \multicolumn{1}{p{0.6cm}|}{ 75.96 74.63} &  \multicolumn{1}{p{0.6cm}|}{62.57  61.43} &  \multicolumn{1}{p{0.6cm}|}{58.59  56.98} &  \multicolumn{1}{p{0.6cm}|}{41.63  35.0} &  \multicolumn{1}{p{0.6cm}|}{54.88 69.29} &  \multicolumn{1}{p{0.6cm}|}{55.93  67.0} &  \multicolumn{1}{p{0.6cm}|}{64.25  64.68} &  \multicolumn{1}{p{0.6cm}|}{60.05  66.51} &  \multicolumn{1}{p{0.6cm}|}{64.54  72.23}   \\

\hline
\hline

{\bf PBoH}& \multicolumn{1}{p{0.6cm}|}{\textcolor{red}{87.19 90.40}} & \multicolumn{1}{p{0.6cm}|}{\textcolor{red}{86.72 86.85}} & \multicolumn{1}{p{0.6cm}|}{\textcolor{red}{86.63 85.48}} & \multicolumn{1}{p{0.6cm}|}{\textcolor{red}{87.39 86.32}} & \multicolumn{1}{p{0.6cm}|}{\textcolor{red}{86.59 87.30}} & \multicolumn{1}{p{0.6cm}|}{\textcolor{red}{86.64 86.14}} & \multicolumn{1}{p{0.6cm}|}{\textcolor{red}{79.48 80.13}} & \multicolumn{1}{p{0.6cm}|}{62.47 61.04} & \multicolumn{1}{p{0.6cm}|}{\textcolor{blue}{61.70 55.83}} & \multicolumn{1}{p{0.6cm}|}{\textcolor{red}{74.19 84.48}} & \multicolumn{1}{p{0.6cm}|}{\textcolor{red}{73.08 81.25}} & \multicolumn{1}{p{0.6cm}|}{\textcolor{red}{89.54 89.62}} & \multicolumn{1}{p{0.6cm}|}{\textcolor{red}{76.54 83.31}} & \multicolumn{1}{p{0.6cm}|}{\textcolor{blue}{71.24 78.33}} \\

\hline
\end{tabular}

\caption{Micro and macro F1 scores reported by Gerbil for 14 datasets and 11 entity linking systems including PBoH. For each dataset and each metric, we highlight in red the best system and in blue the second best system.}
\label{gerbil}
\end{table*}

We now present the experimental evaluation of our method. We first uncover some practical details of our approach. Further, we show  an empirical comparison between PBoH and well known or recent competitive  entity disambiguation systems. We use the Gerbil testing platform~\cite{gerbil} version 1.1.4 with the D2KB setting in which a document together with a fixed set of mentions to be annotated are given as input. We run additional experiments that allow us to compare against more recent approaches, such as ~\cite{houlsby2014scalable} and ~\cite{rel-rw}.

\par Note that in all the experiments we assume that we have access to a set of linkable token spans for each document. In practice this set is obtained by first applying a mention detection approach which is not part of our method. Our main goal is then to annotate each token span with a Wikipedia entity\footnote{\label{note1}In PBoH, we refrain from annotating mentions for which no candidate entity is found according to the procedure described in Section~\ref{ssec:candidates}.}.

\paragraph*{Evaluation metrics}

We quantify the quality of an entity linking system by measuring common metrics such as \textbf{precision}, \textbf{recall} and \textbf{$\mathbf{F_1}$ scores}. 

Let $M^*$ be the ground truth entity annotations associated with a given set of mentions $X$. Note that in all the results reported, mentions that contain NIL or empty ground truth entities are discarded before the evaluation; this decision is taken as well in Gerbil version 1.1.4. Let $M$ be the output annotations of an entity disambiguation system on the same input. Then, our quality metrics are computed as follows:
\begin{itemize}
\item Precision: $P = \frac{\vert M \cap M^* \vert}{\vert M \vert}$
\item Recall: $R = \frac{\vert M \cap M^* \vert}{\vert M^* \vert}$ 
\item $F_1$ score: $F_1 = \frac{2 \cdot P \cdot R}{P + R}$
\end{itemize}

We mostly report results in terms of $F_1$ scores, namely macro-averaged \textbf{F1@MA} (aggregated across documents), and micro-averaged \textbf{F1@MI} (aggregated across mentions). For a fair comparison with Houlsby and Ciaramita~\cite{houlsby2014scalable}, we also report micro-recall \textbf{R@MI} and macro-recall \textbf{R@MA} on the AIDA datasets.

Note that, in our case, the precision and recall are not necessarily identical since a method may not consider annotating certain mentions \footref{note1}.

\begin{table}
\begin{tabular}{@{}|l||c|c||c|c|@{}}
\cline{2-5}
\multicolumn{1}{c||}{} & \multicolumn{4}{c|}{Datasets}   \\
\cline{2-5}
\multicolumn{1}{c||}{} & \multicolumn{2}{c|}{AIDA test A} & \multicolumn{2}{c|}{AIDA test B} \\

\hline
{\bf Systems} & {\bf R@MI} & {\bf R@MA} & {\bf R@MI} & {\bf R@MA} \\
\hline
\hline
LocalMention & 69.73 & 69.30 & 67.98 & 72.75 \\
\hline

TagMe reimpl. & 76.89 & 74.57 & 78.64 & 78.21\\
\hline
AIDA & 79.29 & 77.00 & 82.54 & 81.66\\
\hline
S \& Y & - & 84.22 & - & - \\
\hline
Houlsby et al. & 79.65 & 76.61 & 84.89 & 83.51 \\
\hline
\hline
{\bf PBoH} & {\textcolor{red}{85.70}} & {\textcolor{red}{85.26}} & {\textcolor{red}{87.61}} & {\textcolor{red}{86.44}}\\

\hline
\end{tabular}
\caption{AIDA test-a and AIDA test-b datasets results.}
\label{AIDA:results}
\end{table}

\begin{table*}
\scriptsize
\centering
\begin{tabular}{@{}|l||c|c||c|c||c|c|@{}}

\cline{2-7}
\multicolumn{1}{c||}{} & \multicolumn{6}{c|}{Datasets}   \\
\cline{2-7}
\multicolumn{1}{c||}{} & \multicolumn{2}{c|}{new MSNBC} & \multicolumn{2}{c||}{new AQUAINT} & \multicolumn{2}{c|}{new ACE2004}  \\

\hline
{\bf Systems} & {\bf F1@MI} & {\bf F1@MA} & {\bf F1@MI} & {\bf F1@MA}  & {\bf F1@MI} & {\bf F1@MA} \\
\hline
\hline

LocalMention & 73.64 & 77.71 & 87.33 & 86.80 & 84.75 & 85.70 \\
\hline
\hline
Cucerzan & 88.34& 87.76 & 78.67& 78.22 & 79.30 &78.22 \\
\hline
M \& W &78.43& 80.37& 85.13& 84.84& 81.29& 84.25\\
\hline
Han et al. & 88.46& 87.93& 79.46& 78.80& 73.48& 66.80\\
\hline
AIDA & 78.81& 76.26& 56.47& 56.46& 80.49& 84.13\\
\hline
GLOW & 75.37& 77.33& 83.14& 82.97& 81.91& 83.18\\
\hline
RI & 90.22& 90.87& 87.72& 87.74& 86.60& 87.13\\
\hline
REL-RW & {\textcolor{red}{91.37}} & {\textcolor{red}{91.73}} & {\textcolor{red}{90.74}} & {\textcolor{red}{90.58}} & {\textcolor{blue}{87.68}} & {\textcolor{red}{89.23}} \\
\hline
\hline

{\bf PBoH}& {\textcolor{blue}{91.06}} & {\textcolor{blue}{91.19}} & {\textcolor{blue}{89.27}} & {\textcolor{blue}{88.94}} & {\textcolor{red}{88.71}} & {\textcolor{blue}{88.46}} \\

\hline
\end{tabular}

\caption{Results on the newer versions of the MSNBC, AQUAINT and ACE04 datasets.}
\label{ACE:results}
\end{table*}

\paragraph*{Pseudo-likelihood training}
 
We briefly mention some of the practical issues that we encounter with the likelihood maximization described in Section~\ref{ssec:pseudolikelihood}. From the practical perspective, for each mention $m$, we only considered the set of parameters $\rho_{m,e}$ limited to the top 64 candidate entities $e$ per mention, ranked by $\hat{p}(e|m)$ . Additionally, we restricted the set $\lambda_{e,e'}$ to entity pairs $(e,e')$ that co-occurred together in at least 7 documents throughout the Wikipedia corpus. In total, a set of 26 millions $\rho$ and 39 millions $\lambda$ parameters were learned using the previously described procedure. Note that the universe of all Wikipedia entities is of size $\sim$ 4 million.

For the SGD procedure, we tried different initializations of these parameters, including $\rho_{m,e} = \log\ p(e|m), \lambda_{e,e'} = 0$, as well as the parameters given by Eq.~\eqref{eq:lambda-bethe}. However, in all cases, the accuracy gain on a sample of 1000 Wikipedia test pages was small or negligible compared to the LocalMention baseline (described below). One reason is the inherent sparsity of the data: the parameters associated with the long tail of infrequent entity pairs are updated rarely and expected to be defective at the end of the SGD procedure. However, these scattered pairs are crucial for the effectiveness and coverage of the entity disambiguation system. To overcome this problem, we refined our model as described in Section~\ref{ssec:unnormalized} and subsequent sections. \\

\paragraph*{PBoH training details}
Wi\-ki\-pe\-dia itself is a valuable resource for entity linking since each internal hyperlink can be considered as the ground truth annotation for the respective anchor text. In our system, the training is solely done on the entire Wi\-ki\-pe\-dia corpus\footnote{We used the Wikipedia dump from February 2014}. Hyper-parameters are grid-searched such that the micro $F_1$ plus macro $F_1$ scores are maximized over the combined held-out set containing only the \textit{AIDA Test-A} dataset and a Wi\-ki\-pe\-dia validation set consisting of random 1000 pages.  As a preprocessing step in our training procedure, we removed all annotations and hyperlinks that point to non-existing, disambiguation or list Wi\-ki\-pe\-dia pages.

The \textit{PBoH} system used in the experimental comparison is the  model given by Eq.~\eqref{eq:c2} for which grid search of the hyper-parameters suggested using $\zeta = 0.075,\tau = 0.5,\delta=0.5, \xi=0.5$.

\paragraph*{Datasets}

We evaluate our approach on 14 well-known public entity linking datasets built from various sources. Statistics of some of them are shown in Table~\ref{Ta:datastats}, and their descriptions are provided below. For information on the other datasets used only in the Gerbil experiments, refer to \cite{gerbil}.
\begin{itemize}
\item The \textit{CoNLL-AIDA} dataset is an entity annotated corpus of Reuters news documents introduced by Hoffart et al.~\cite{aida}. It is much larger than most of the other existing EL datasets, making it an excellent evaluation target. The data is divided in three parts: Train (not used in our current setting for training, but only in the Gerbil evaluation), Test-A (used for validation) and Test-B (used for blind evaluation). Similar to Houlsby and Ciaramita~\cite{houlsby2014scalable} and others, we report results also on the validation set Test-A.
\item The \textit{AQUAINT} dataset introduced by Milne and Witten~\cite{milne2008learning} contains documents from a news corpus from the Xinhua News Service, the New York Times and the Associated Press.
\item \textit{MSNBC}~\cite{cucerzan2007large} - a dataset of news documents that includes many mentions which do not easily map to Wikipedia titles because of their rare surface forms or distinctive lexicalization.
\item The \textit{ACE04} dataset~\cite{ratinov2011local} is a subset of ACE2004 Coreference documents annotated using Amazon Mechanical Turk. Note that the ACE04 dataset contains mentions that are annotated with NIL entities, meaning that no proper Wikipedia entity was found. Following common practice, we removed all the mentions corresponding to these NIL entities prior to our evaluation.
\end{itemize}

Note that the Gerbil platform uses an old version of the AQUAINT, MSNBC and ACE04 datasets that contain some no-longer existing Wikipedia entities. A new cleaned version of these sets\footnote{\url{http://www.cs.ualberta.ca/~denilson/data/deos14_ualberta_experiments.tgz}} was released by Guo \& Barbosa~\cite{rel-rw}. We report results for the new cleaned datasets in Table~\ref{ACE:results}, while Table~\ref{gerbil} contains results for the old versions currently used by Gerbil.

\begin{table}
\begin{tabular}{@{}|p{1.5cm}||c|c|c|c|c|@{}}
\cline{2-6}
\multicolumn{1}{c||}{} & \multicolumn{5}{c|}{Datasets}   \\
\cline{2-6}
\multicolumn{1}{c||}{} & \multicolumn{1}{p{1cm}|}{\tiny{AIDA test A}} & \multicolumn{1}{p{1cm}|}{\tiny{AIDA test B}} & \multicolumn{1}{c|}{\tiny{MSNBC}} & \multicolumn{1}{c|}{\tiny{AQUAINT}} & \multicolumn{1}{c|}{\tiny{ACE04}} \\
\hline
\hline
\small{Avg. num mentions per doc} & 22.18 & 19.41 & 32.8 & 14.54 & 7.34 \\
\hline
\small{Conv. rate} & 100\% & 99.56\% & 100\% & 100\% & 100\% \\
\hline
\small{Avg. running time (ms/doc)} & 445.56 & 203.66 & 371.65 & 40.42 & 10.88 \\
\hline
\small{Avg. num. rounds} & 2.86 & 2.83 & 3.0 & 2.56 & 2.25\\
\hline
\end{tabular}
\caption{Loopy belief propagation statistics. Average running time, number of rounds and convergence rate of our inference procedure are provided.}
\label{Ta:running time}
\end{table}

\paragraph*{Systems}

For comparison, we selected a broad range of competitor systems from the vast literature in this field. The Gerbil platform already integrates the methods of 
Agdistis~\cite{agdistis}, Babelfy~\cite{babelfy}, DBpedia Spotlight~\cite{dbpediaspotlight}, Dexter~\cite{dexter}, Kea~\cite{kea},  Nerd-ML~\cite{nerd-ml}, Tagme2~\cite{tagme2}, WAT~\cite{wat}, Wikipedia Miner~\cite{milne2008learning} and Illinois Wikifier~\cite{ratinov2011local}. We furthermore compare against Cucerzan~\cite{cucerzan2007large} -- the first collective EL system that uses optimization techniques, M\& W~\cite{milne2008learning}-- a popular machine learning approach, Han et al.~\cite{han2011collective} -- a graph based disambiguation system that uses random walks for joint disambiguation, AIDA~\cite{aida} -- a performant graph based approach, GLOW~\cite{ratinov2011local} -- a system that uses local and global context to perform joint entity disambiguation, RI~\cite{cheng2013relational} -- an approach using relational inference for mention disambiguation, and REL-RW~\cite{rel-rw}, a recent system that iteratively solves mentions relying on an online updating random walk model. In addition, on the AIDA datasets we also compare against S\& Y~\cite{sil2013re} -- an apparatus for combining the NER and EL tasks, and Houlsby et al.~\cite{houlsby2014scalable} -- a topic modelling LDA-based approach for EL.

To empirically assess the accuracy gain introduced by each incremental step of our approach, we ran experiments on several of our method's components, individually: \textbf{LocalMention} -- links mentions to entities solely based on the token span statistics, i.e., $e^* = \argmax_e \hat{p}(e|m)$;  \textbf{Unnorm} -- uses the unnormalized mention-entity model described in Section ~\ref{ssec:unnormalized}; \textbf{Rescaled} -- relies on the rescaled model presented in Section ~\ref{ssec:scaling-len}; \textbf{LocalContext} -- disambiguates an entity based on the mention and the local context probability given by Equation~\eqref{eq:local-all}, i.e., $e^* = \argmax_e p(e|m,c)$. Note that \textbf{Unnorm}, \textbf{Rescaled} and \textbf{PBoH} use the loopy belief propagation procedure for inference.

\subsection{Results}

Results of the experiments run on the Gerbil platform are shown in Table~\ref{gerbil}. Detailed results are also provided\footnote{The PBoH Gerbil experiment is available at \url{http://gerbil.aksw.org/gerbil/experiment?id=201510160025}.}\footnote{The detailed Gerbil results of the baseline systems can be accessed at \url{http://gerbil.aksw.org/gerbil/experiment?id=201510160026}}. We obtain the highest performance on 11 datasets and the second highest performance on 2 datasets, showing the effectiveness of our method.

Other results are presented in Table~\ref{AIDA:results} and Table~\ref{ACE:results}. The highest accuracy for the cleaned version of AQUAINT, MSNBC and ACE04 was previously reported by Guo \& Barbosa~\cite{rel-rw}, while Houlsby et al.~\cite{houlsby2014scalable}  dominate the AIDA datasets. Note that the performance of the baseline systems shown in these two tables is taken from ~\cite{rel-rw} and ~\cite{houlsby2014scalable}. 

All these methods are tested in the setting where a fixed set of mentions is given as input, without requiring the mention detection step.

\paragraph*{Discussion}

Several observations are worth noting here. First,
the simple \textit{LocalMention} component alone outperforms many EL
systems. However, our experimental results show that \textit{PBoH}
consistently beats \textit{LocalMention} on all the
datasets. Second, \textit{PBoH} produces state-of-the-art results on
both development (Test-A) and blind evaluation (Test-B) parts of the AIDA
dataset. Third, on the AQUAINT, MSNBC and ACE04
datasets, \textit{PBoH} outperforms all but one of the presented EL
systems and is competitive with the state-of-art approaches. The
method whose performance is closer to ours is REL-RW~\cite{rel-rw}
whose average F1 score is only slightly higher than ours (\textit{+0.6
on average}). However, there are significant advantages of
our method that make it easier to use for practitioners. First, our
approach is conceptually simpler and only requires sufficient
statistics computed from Wikipedia. Second, \textit{PBoH} shows a
superior computational complexity manifested in significantly lower
run times (Table~\ref{Ta:running time}), making it a good fit for
large-scale real-time entity linking systems; this is not the case for
REL-RW qualified as ``time consuming'' by its
authors. Third, the number of entities in the underlying
graph, and thus the required memory, is significantly lower
for \textit{PBoH} (see statistics provided in Table~\ref{guo}).

\begin{table}
\begin{tabular}{@{}|p{2.5cm}||c|c|c|@{}}
\cline{2-4}
\multicolumn{1}{p{1.5cm}||}{} & \multicolumn{3}{c|}{Datasets}   \\
\cline{2-4}
\multicolumn{1}{p{1.5cm}||}{} & \multicolumn{1}{c|}{MSNBC } & \multicolumn{1}{c|}{AQUAINT} & \multicolumn{1}{c|}{ACE2004}  \\
\hline
{\bf Avg \# mentions per doc} & 36.95 & 14.54 &  8.68 \\
\hline
{\bf Systems} & \# entities & \# entities &  \# entities \\
\hline
PBoH & 247.19 & 95.38 & 66.66\\
\hline
REL-RW & 382773.6 & 242443.1 & 256235.49 \\
\hline
\end{tabular}
\caption{Average number of entities that appear in the graph built by PBoH and by REL-RW}
\label{guo}
\end{table}

\begin{table}
\begin{tabular}{@{}|l||c|c||c|c|@{}}
\cline{2-5}
\multicolumn{1}{c||}{} & \multicolumn{4}{c|}{Datasets}   \\
\cline{2-5}
\multicolumn{1}{c||}{} & \multicolumn{2}{c|}{AIDA test A} & \multicolumn{2}{c|}{AIDA test B} \\

\hline
{\bf Systems} & {\bf R@MI} & {\bf R@MA} & {\bf R@MI} & {\bf R@MA} \\
\hline
\hline
LocalMention & 69.73 & 69.30 & 67.98 & 72.75 \\
\hline

Unnorm & 69.77 & 69.95 & 75.87 & 75.12 \\
\hline
Rescaled & 75.09 & 74.25 & 74.76 & 78.28\\
\hline
LocalContext & 82.50 & 81.56 &  85.46 & 84.08 \\
\hline
{\bf PBoH} & {\bf 85.53 } & {\bf 85.09} & {\bf 87.51} & {\bf 86.39}\\
\hline
\end{tabular}
\caption{Accuracy gains of individual \textit{PBoH} components.}
\label{Ta:incremental}
\end{table}

\subsubsection*{Incremental accuracy gains}

To give further insight to our method, Table~\ref{Ta:incremental} provides an overview of the contribution brought step by step by each incremental component of the \textit{Full PBoH} system. It can be noted that \textit{PBoH} performs best, outranking all its individual components.

\subsubsection*{Reproducibility of the experiments}
Our experiments are easily reproducible using the details provided in this paper. Our learning procedure is only based on statistics coming from the set of Wikipedia webpages. As a consequence, one can implement a real-time highly accurate entity disambiguation system solely based on the details described in this paper. 

Our code is publicly available at : \url{https://github.com/dalab/pboh-entity-linking}

\section{Conclusion}\label{sec:conclusion}

In this paper, we described a light-weight graphical model for entity linking  via approximate inference. Our method employs simple sufficient statistics that rely on three sources of information: First, a probabilistic name to entity map $\hat p(e|m)$ derived from a large corpus of hyperlinks; second,  observational data about the pairwise co-occurrence of entities within documents from a Web collection; third, entity - contextual words statistics. Our experiments based on a number of popular entity linking benchmarking collections show improved performance as compared to several well-known or recent systems.\\

There are several promising directions of future work. Currently, our model considers only pairwise potentials. In the future, it would be interesting to investigate the use of higher-order potentials and submodular optimization in an entity linking pipeline, thus allowing us to capture the interplay between entire groups of entity candidates (e.g., through the use of entity categories). Additionally, we will further enrich our probabilistic model with statistics from new sources of information. We expect some of the performance gains that other papers report from using entity categories or semantic relations to be additive with regard to our system's current accuracy.

\small
\bibliographystyle{abbrv}
\bibliography{references}
\normalsize

\end{document}